\documentclass[10pt,twocolumn,letterpaper]{article}

\usepackage{iccv}
\usepackage{times}
\usepackage{epsfig}
\usepackage{graphicx}
\usepackage{amsmath}
\usepackage{amssymb}

\usepackage{enumitem}
\usepackage{subcaption}
\usepackage[accsupp]{axessibility}

\usepackage[pagebackref=true,breaklinks=true,letterpaper=true,colorlinks,bookmarks=false]{hyperref}

\iccvfinalcopy 

\def\iccvPaperID{10436} 

\ificcvfinal\fi

\begin{document}

\title{Beyond Road Extraction: A Dataset for Map Update using Aerial Images}

\author{Favyen Bastani \\
MIT CSAIL \\
{\tt\small favyen@csail.mit.edu}
\and
Sam Madden \\
MIT CSAIL \\
{\tt\small madden@csail.mit.edu}}

\maketitle
\ificcvfinal\thispagestyle{empty}\fi

\begin{abstract}
The increasing availability of satellite and aerial imagery has sparked substantial interest in automatically updating street maps by processing aerial images. Until now, the community has largely focused on road extraction, where road networks are inferred from scratch from an aerial image. However, given that relatively high-quality maps exist in most parts of the world, in practice, inference approaches must be applied to update existing maps rather than infer new ones. With recent road extraction methods showing high accuracy, we argue that it is time to transition to the more practical \emph{map update} task, where an existing map is updated by adding, removing, and shifting roads, without introducing errors in parts of the existing map that remain up-to-date. In this paper, we develop a new dataset called MUNO21 for the map update task, and show that it poses several new and interesting research challenges. We evaluate several state-of-the-art road extraction methods on MUNO21, and find that substantial further improvements in accuracy will be needed to realize automatic map update.
\end{abstract}

\section{Introduction}

Maintaining street map datasets is an extraordinarily labor-intensive and expensive process, with several companies today spending hundreds of millions of dollars annually on the task~\cite{appleinvestmaps,uberinvestmaps}. The cost of updating maps, coupled with the increasing availability of aerial and satellite imagery, has sparked much interest in automatically processing aerial images to update maps~\cite{roadtracer,batra2019improved,cheng2017automatic,sat2graph,li2019topological,mattyus2017deeproadmapper,mnih2010learning,mosinska2019joint,sun2019leveraging,tan2020vecroad,wegner2015road,yang2019road,zhou2018d}. To date, most proposed methods have focused on road extraction, where methods infer all roads in an aerial image. The inferred road network is generally represented as a spatial network, i.e., a graph where vertices are assigned longitude-latitude coordinates and edges correspond to detected roads. Road extraction methods are typically evaluated by comparing inferred road networks with a ground truth road network (either hand-labeled in the aerial image or taken from a map dataset).

However, inferred road networks are not directly useful: instead, in practice, inference approaches must identify positions where existing maps do not reflect the current physical road network, and update the map at those positions --- overwriting the existing map would not be sensible since it contains street names, speed limits, tunnels, and other crucial features not visible from an aerial view.

\begin{figure}[t]
\begin{subfigure}{\linewidth}
    \centering
    \setlength{\tabcolsep}{1pt}
    \begin{tabular}{ccc}
        2012 & 2019 & Map \\
    	\includegraphics[width=0.31\linewidth]{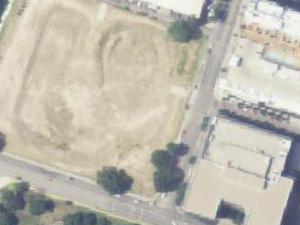} &
    	\includegraphics[width=0.31\linewidth]{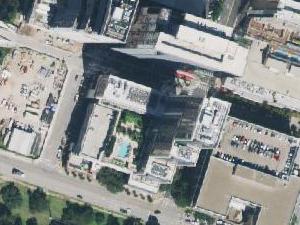} &
    	\includegraphics[width=0.31\linewidth]{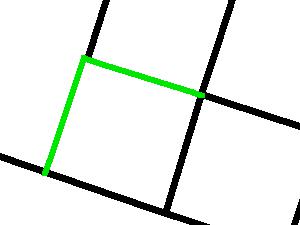} \\
	\end{tabular}
	\vspace{-8pt}
	\caption{Scenario tagged Constructed: the construction of a new road is apparent in the aerial images.}
\end{subfigure}
\begin{subfigure}{\linewidth}
    \centering
	\includegraphics[width=0.31\linewidth]{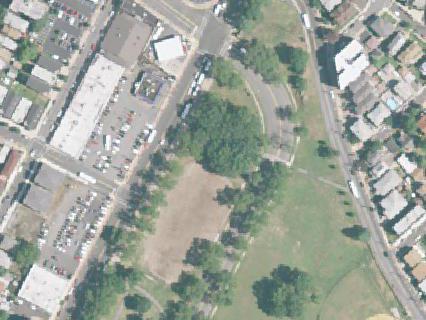}
	\includegraphics[width=0.31\linewidth]{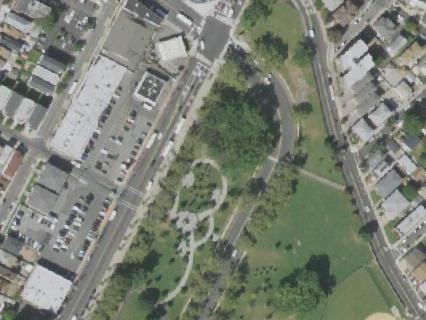}
	\includegraphics[width=0.31\linewidth]{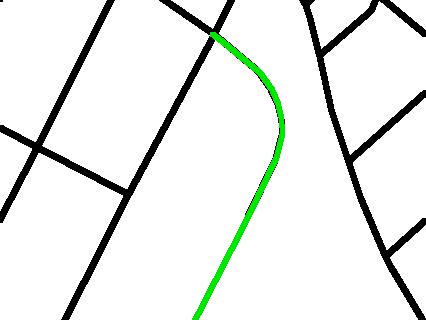}
	\vspace{-4pt}
	\caption{Scenario tagged Was-Missing: a road that was not recently constructed is missing from the map.}
\end{subfigure}
\begin{subfigure}{\linewidth}
    \centering
	\includegraphics[width=0.31\linewidth]{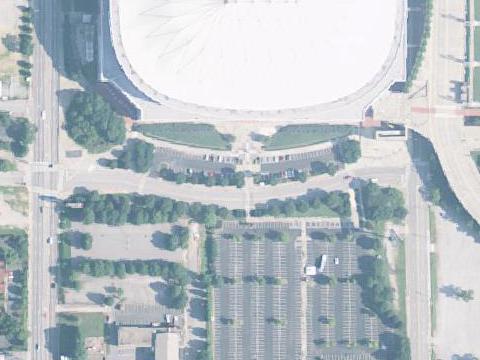}
	\includegraphics[width=0.31\linewidth]{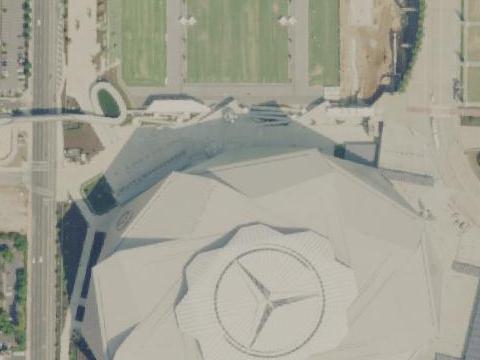}
	\includegraphics[width=0.31\linewidth]{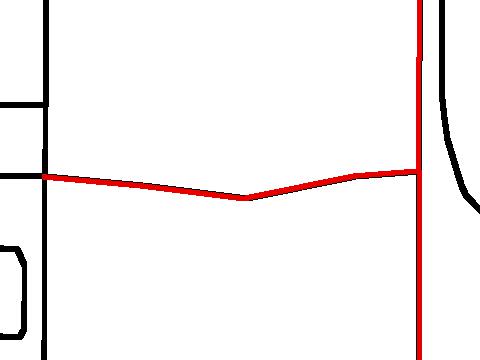}
	\vspace{-4pt}
	\caption{Scenario tagged Deconstructed: road was physically removed.}
\end{subfigure}
\begin{subfigure}{\linewidth}
    \centering
	\includegraphics[width=0.31\linewidth]{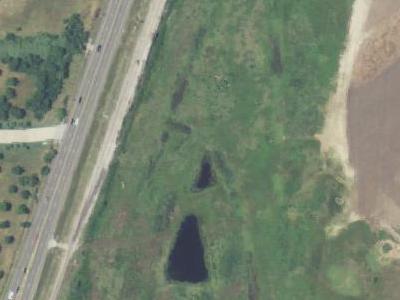}
	\includegraphics[width=0.31\linewidth]{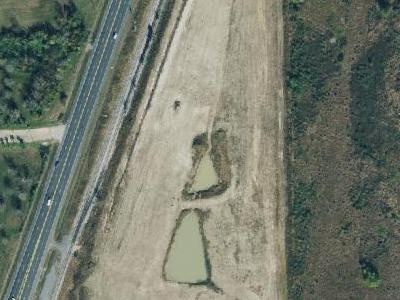}
	\includegraphics[width=0.31\linewidth]{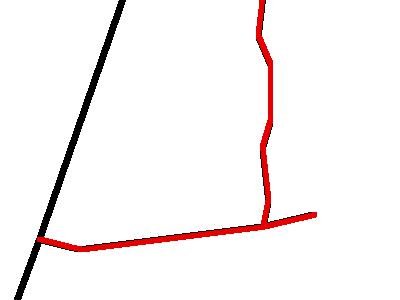}
	\vspace{-4pt}
	\caption{Scenario tagged Was-Incorrect: a road is in the map, but the road never existed in the physical road network.}
\end{subfigure}
\begin{subfigure}{\linewidth}
    \centering
	\includegraphics[width=0.31\linewidth]{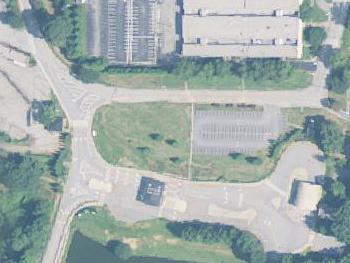}
	\includegraphics[width=0.31\linewidth]{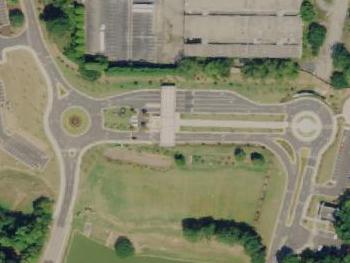}
	\includegraphics[width=0.31\linewidth]{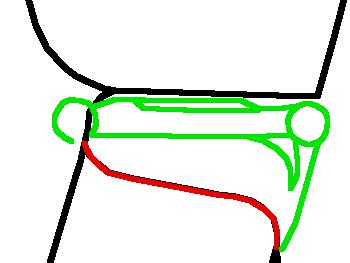}
	\vspace{-4pt}
	\caption{Scenario with multiple tags due to a road topology change.}
\end{subfigure}
\caption{Example map update scenarios in MUNO21. In map update, a method inputs an aerial image time series and an existing map (black), and produces an updated map. We show older images on the left and newer images on the right. Green segments are not yet in the map but should be added; red segments are in the map but should be removed.}
\label{fig:dataset}
\end{figure}

A naive approach for updating maps using existing road extraction methods is to fuse inferred roads into the existing map: we can first remove redundant inferred roads that duplicate roads already present in the map, and integrate the remaining inferred roads. However, this approach has several shortcomings: (a) connections between inferred roads and the existing map may not be inferred accurately; (b) roads in the existing map that have been physically removed or were simply incorrect will remain in the map; and (c) areas where the road topology has changed (e.g., a junction modified into a roundabout) will not be updated correctly.

Thus, map update is not only more directly practical than road extraction, but also poses several new and interesting research challenges. We argue that, with state-of-the-art methods reporting high accuracy scores on road extraction (between 75\% and 85\% across several metrics and datasets~\cite{sat2graph,tan2020vecroad}), it is time for a shift in focus from road extraction to the more general and complex map update task.

In this paper, we develop a large-scale and comprehensive dataset for the map update task. We call our dataset MUNO21\footnote{The MUNO21 dataset and code are available at \url{https://favyen.com/muno21/}.}, for MapUpdate-NAIP/OSM, since we use NAIP aerial imagery\footnote{The National Agriculture Imagery Program (NAIP) provides public domain 60-100 cm/pixel aerial imagery over the continental US.} and OpenStreetMap (OSM) data. MUNO21 spans regions totaling over 6000 sq km in area, spread over 21 cities across the United States. It includes time series of aerial images and map data to capture the evolution of both the physical road network and real street maps over time --- we collect NAIP aerial images at each of four years over the eight-year timespan from 2012--2019, and OSM extracts from each year during the same timespan. We focus on the US in part due to the availability of public domain aerial imagery, but also because the US is a good test case for map update: OpenStreetMap has good coverage and the road network includes complex topology, and at the same time, the map undergoes frequent changes: 30K km of roads are constructed each year\footnote{``Public Road Mielage'', FHWA, \url{https://www.fhwa.dot.gov/policyinformation/statistics/2013/vmt422c.cfm}}, and 50K km of roads are added each month to OSM in the US\footnote{``Total Navigable Roads per Month in US'', \url{https://metrics.improveosm.org/}}.

To evaluate map update methods, we could randomly prune some fraction of roads from the most recent OSM data, and compute how accurately methods can recover the pruned roads. However, this would fail to capture several of the challenges we have discussed above, such as deleting roads from the map that were physically removed. Additionally, the randomly pruned roads would not correspond to the evolution of the physical road network apparent in the aerial image time series.

Instead, we incorporate 514 real \emph{map update scenarios} spanning 265 sq km into MUNO21 by identifying regions of OSM that have changed in non-negligible ways between 2012--2019. We show example scenarios in Figure \ref{fig:dataset}.
Each scenario consists of actual changes made to OSM, and so corresponds directly to real updates that a map update method would need to make in practice. In each scenario, we apply map update methods to reproduce the changes made to OSM: a method inputs a pre-change map and the aerial image time series, and is evaluated in terms of how accurately it reproduces the post-change map. To identify these scenarios, we develop a semi-automated annotation approach where we first automatically compare OSM data across consecutive years to identify a large number of candidate changes, and then manually select a subset of these changes where the new map corresponds to the most recent aerial image, and where the changes are unambiguous. 

We divide the dataset into 10 cities for training and 11 cities for testing. In addition to 3000 sq km of aerial imagery and road network pairs, the training set includes 326 of the 514 map update scenarios. In these scenarios, when a change corresponds to a physical construction event that is apparent in the aerial image time series, we additionally hand-label the year when the construction is first visible, and the year when it is visibly completed. (In other scenarios, the changes are already reflected in the oldest aerial image, even though the changes are made to the map much later, e.g., due to older construction or spurious roads in the dataset.) This training data is especially useful for improving accuracy on capturing updates following construction activities, since the construction year annotations enable training methods that compare aerial images over time to identify physically added and removed roads.

We evaluate five state-of-the-art road extraction and map update methods that have publicly available implementations on MUNO21. We also discuss and evaluate extensions of existing methods for map update that leverage the additional time series training data that MUNO21 provides, which we show improves precision from 98\% to 99.8\% at 20\% recall, but only considers changes corresponding to physical construction.
Our evaluation shows that there is substantial room for improvement in accuracy on the map update problem, and we believe MUNO21 will help enable a transition in focus from road extraction to map update.

The remainder of this paper is organized as follows. We first discuss related work in Section \ref{sec:related}. We then detail the MUNO21 dataset in Section \ref{sec:dataset}, and discuss our approach for annotating the dataset, along with proposed evaluation metrics. In Section \ref{sec:methods}, we propose extensions of prior work that leverage new types of training data in MUNO21. Finally, in Section \ref{sec:eval}, we compare state-of-the-art road extraction methods and our proposed extensions on MUNO21.

\section{Related Work} \label{sec:related}

\noindent
\textbf{Road Extraction.} Early work in automatic road extraction proposed leveraging crowdsourced GPS trajectories to infer road networks~\cite{biagioni2012inferring,buchin2020improved,cao2009gps,chen2016city,davies2006scalable,he2018roadrunner,prabowo2019coltrane,shan2015cobweb,stanojevic2018robust}. These methods were inspired by the widespread use of GPS data in manual map curation; in fact, OpenStreetMap continues to maintain a database of GPS trajectories contributed by the community that users can leverage when editing the map. However, GPS trajectories have poor coverage outside urban centers. With the increasing availability of satellite and aerial imagery with global coverage, interest has shifted towards automatically extracting roads from aerial images. Many methods propose segmenting aerial images for roads using CNNs~\cite{batra2019improved,mattyus2017deeproadmapper,mosinska2019joint,yang2019road,zhou2018d}. To extract road networks from segmentation probabilities, a combination of binary thresholding, morphological thinning, and line following is typically employed~\cite{cheng2017automatic}. Other techniques propose extracting road networks through a point-by-point search process: RoadTracer and PolyMapper propose training an agent with supervised learning to walk along roads visible in an aerial image, and constructing a road network along the agent's path~\cite{roadtracer,li2019topological}. VecRoad extends this approach with a flexible step size and joint learning tasks~\cite{tan2020vecroad}, and Neural Turtle Graphics extends it with a sequential generative model~\cite{chu2019neural}. Another recent technique, Sat2Graph, proposes a one-shot road extraction process where a CNN directly predicts the positions of road network vertices and edges~\cite{sat2graph}.

\smallskip
\noindent
\textbf{Map Update.} A few methods have considered updating maps using either GPS trajectories or aerial images. However, these methods have focused on a narrow subset of the map update problem (e.g., only adding roads), and evaluation metrics have largely been qualitative due to the lack of a large-scale and realistic dataset. CrowdAtlas~\cite{wang2013crowdatlas} and COBWEB~\cite{shan2015cobweb} propose matching GPS trajectories to the current map, and aggregating unmatched trajectories to form road proposals.
MAiD proposes an extension of point-by-point search techniques for inferring roads in aerial images that are not covered by an existing map, along with connections between new roads and existing roads~\cite{maid}.

\smallskip
\noindent
\textbf{Datasets.} Several datasets have been curated for road extraction. DeepGlobe 2018 incorporates 2,220 sq km of 50 cm/pixel Maxar satellite imagery and corresponding road annotations in regions of Thailand, Indonesia, and India~\cite{deepglobe}. SpaceNet 3 incorporates 3000 sq km of satellite imagery and road centerlines in Las Vegas, Paris, Shanghai, and Khartoum~\cite{spacenet}. To our knowledge, MUNO21 is the first dataset for the map update task that includes pairs of pre-change maps and recent aerial images, along with ground truth post-change maps. Additionally, in contrast to prior datasets, MUNO21 includes a time series of aerial images rather than images at a single point in time; analyzing multiple aerial images of the same location over time is especially useful for detecting constructed roads, deconstructed roads, and road network topology changes.

\section{MUNO21} \label{sec:dataset}

The MUNO21 dataset incorporates aerial imagery from the National Agriculture Imagery Program (NAIP) and street map data from OpenStreetMap (OSM) covering a total area of 6,052 sq km across 21 cities in the United States. In each city, we use historical bi-annual NAIP imagery and the OSM history dump to extract aerial images at each of four years between 2012--2019, and map data at every year.

The core component of MUNO21 is a set of 514 map update scenarios spanning 265 sq km in which substantial changes to the road network have been made in OSM during the eight-year period, along with an additional 780 no-change scenarios where neither the road network in OSM nor the physical road network apparent in the aerial imagery have changed. Each scenario specifies a bounding box $(x, y, w, h)$ defining the window that has changed, the pre-change map $G$ taken from OSM before the map was changed, and the post-change map $G^{*}$. We split the dataset into a training set with 10 cities and 726 corresponding scenarios, and a test set with 11 cities and 568 corresponding scenarios. To evaluate a map update method on a scenario, we provide the method as input the pre-change map $G$, and the aerial image time series at the scenario window. The method should output a new road network $\hat{G}$.
In scenarios with change, we measure how well the method captures change by comparing the output map $\hat{G}$ with the ground truth map $G^{*}$. In scenarios with no change, the method should not make any changes to the map; we use the no-change scenarios to measure a method's precision, i.e., the fraction of no-change scenarios where $\hat{G} = G$.

We annotate each scenario in the training set with one or more of five tags:
\begin{itemize}[noitemsep,topsep=0pt]
    \item \textbf{No-Change} indicates neither the map nor the physical roads have changed.
    \item \textbf{Constructed} indicates that roads were added to the map, and that the addition corresponds to a recent physical road network change that is apparent in the aerial imagery. In other words, we should be able to see the pre-change map in an earlier image, and the post-change map in a later image.
    \item \textbf{Was-Missing} indicates that roads were added to the map, but, based on the imagery, those roads were already present in the physical road network as of 2012. Thus, the roads were simply missing from the map.
    \item \textbf{Deconstructed} indicates that roads were removed from the map, and the removal corresponds to a recent physical road network change.
    \item \textbf{Was-Incorrect} indicates that roads were removed from the map, but the roads were not visible in the physical road network as of 2012 --- thus, the roads were either bulldozed prior to 2012, or never existed.
\end{itemize}
Additionally, for scenarios tagged Constructed or Deconstructed, we annotate the year when the change was first visible in an aerial image, and the year when construction activity was first visibly completed. These additional annotations are helpful for training methods that compare aerial images over time; these methods may greatly improve precision, but may only be able to make changes to the map that correspond to changes in the physical road network (rather than incorporating pre-existing missing roads).

In this section, we first detail our semi-automated method for annotating the 1,294 map update scenarios. We then propose evaluation metrics to compare map update methods on MUNO21. Finally, we discuss extending road extraction methods for map update by fusing inferred roads into the existing map.

\subsection{Annotation}

\noindent
\textbf{Collecting Imagery and Map Data.} We first identify a total of 39 tiles for the dataset, each of approximately 155 sq km in size (over 6000 sq km total), with one or two tiles in each of the 21 cities depending on the size of the city. We select the tiles to minimize the undeveloped area that contains no roads, e.g. due to bodies of water, forests, or desert. We download the OpenStreetMap history dump for the US from Geofabrik\footnote{https://download.geofabrik.de/north-america/us.html}, and use osmium~\cite{osmium} to extract street map data at each year and at each tile. We use Google EarthEngine to fetch NAIP images at each tile, and to merge and rectify the images under the Web-Mercator projection. We extract the NAIP images at 1 meter/pixel resolution.

\smallskip
\noindent
\textbf{Finding Changes.}
Next, we employ an automated candidate generation approach to identify candidate roads that changed in OSM between 2012--2019.
In our approach, we extract the OSM map at each year, and compare the extracted maps between consecutive years. When comparing a pair of old and new maps $G_\text{old}$ and $G_\text{new}$, we are interested both in roads present in $G_\text{new}$ that are a non-negligible distance away from any road in $G_\text{old}$ (added roads), and roads in $G_\text{old}$ that are far from roads in $G_\text{new}$ (removed roads).

Then, we want to design a method $f(G_1, G_2)$ that, given graphs $G_1$ and $G_2$, identifies road segments in $G_2$ that do not appear in $G_1$. We can then apply the method in both directions, computing $f(G_\text{old}, G_\text{new})$ to find added roads and computing $f(G_\text{new}, G_\text{old})$ to find removed roads. We employ a simple image-based method to implement $f$: (1) we render lines along each road in $G_1$; (2) we dilate the image by $D_\text{big} = 32$ m along both axes, where $D_\text{big}$ is a distance threshold --- we only want to select roads in $G_2$ that are at least $D_\text{big}$ away from any road in $G_1$; and (3), we iterate over each road in $G_2$, and check whether some pixel along the road is not set in the dilated image --- if so, the road is an added or removed road.

We filter the detected added and removed roads with two additional constraints. First, we discard connected components of detected roads that have a total length shorter than a minimum length threshold $L = 70$ m. Second, we discard roads that are labeled as parking, driveway, or service road in OSM.
These constraints help focus automated candidate generation on public streets: while other road types such as driveways are important in a map, we opt to focus on public streets because they are less ambiguous (e.g., driveway vs wide footpath).
We call each remaining connected component of roads a candidate change.

\smallskip
\noindent
\textbf{Clustering Candidates.}
Before proceeding to manual annotation, we group candidate changes that are close in space and time into clusters. This step is motivated by two issues. First, for simplicity, we want map update scenarios to specify a region of interest as a bounding box, rather than a more complex shape; however, the bounding box around one change may contain other changes. Second, if changes are made to the map during several consecutive years, the map is more likely to correspond to the physical road network before and after these changes than when it is in an intermediate state.

Thus, we apply a simple spatio-temporal agglomerative clustering procedure over the changes. We initialize one cluster for each change. We say that a pair of clusters $c_1$ and $c_2$ are mergeable if both (1) the distance between the bounding boxes of $c_1$ and $c_2$ is less than a threshold (128 m); and (2) the time windows of $c_1$ and $c_2$ overlap. We then repeatedly iterate over clusters and combine mergeable clusters until no two clusters are mergeable.

\smallskip
\noindent
\textbf{Manual Annotation.}
Although automatic candidate generation and clustering produces several clusters where OSM has changed, many of the changes may still correspond to driveways, service roads, parking, and other features are highly ambiguous and may not be visible in aerial imagery. Although we have discarded changes corresponding to roads labeled with these types in OSM during candidate generation, incorrect road type labels are common especially in recently added roads. Furthermore, we find that some clusters contain roads that are still under construction and not fully visible in aerial imagery yet.

Thus, we develop a hand-labeling tool to select a subset of the clusters where the changes affect public streets,
and where the changes are visible in the most recent aerial image. We also use the annotation tool to tag each change as Constructed, Was-Missing, Deconstructed, and/or Was-Incorrect, along with the years of the construction activity for changes tagged Constructed or Deconstructed.

Both in the dataset and in the tool, the post-change map is split into two graphs: a public-streets-only graph $G^{*}$ that excludes parking, driveway, and service roads, and an extra-roads graph $G^{*}_\text{extra}$ that includes those roads as well as other ``ways'' (edges) in OSM such as footpaths. This is because a method should not be penalized for failing to produce a road in $G^{*}_\text{extra}$ (since these roads are ambiguous), but should also not be penalized for producing it. In the next section, we will discuss how we handle $G^{*}_\text{extra}$ when evaluating map update methods. For now, in order to ensure an accurate evaluation later, it is important that $G^{*}_\text{extra}$ contain all roads visible in the aerial image, including roads that may be missing from the most recent OSM data. Thus, during manual annotation, we trace additional edges in $G^{*}_\text{extra}$ to ensure that missing roads are covered.

\smallskip
\noindent
\textbf{No-Change Scenarios.} We use a similar generation approach to find No-Change scenarios, where we search for windows in which every road in $G_\text{old}$ is at most $D_\text{small} = 8$ m from a road in $G_\text{new}$, and vice versa. We do not apply clustering; instead, we ensure that computed windows do not overlap. We perform manual annotation over the automatically selected windows to verify that the physical road network has not changed, and to ensure $G^{*}_\text{extra}$ is complete.

\smallskip
\noindent
\textbf{Summary.} Automatic candidate generation and clustering produces 1,812 clusters. After hand-labeling, we retain 514 clusters to use as map update scenarios, of which 164 are tagged Constructed, 259 are tagged Was-Missing, 48 are tagged Deconstructed, and 100 are tagged Was-Incorrect (some scenarios have multiple tags). We also derive 780 No-Change scenarios. The resulting 1,294 scenarios cover 552 sq km and include 948 km of changed roads.

\subsection{Metrics}

We evaluate map update methods on the 568 windows in the test set, of which 188 contain change and 380 are tagged No-Change. On each scenario, a method inputs the pre-change road network $G$ and the aerial image time series, and should output an updated road network $\hat{G}$ that is similar to the post-change road network (ground truth) $G^{*}$.

We compare map update methods in terms of their precision-recall curves. We find that methods generally have a key parameter that provides a tradeoff between precision (avoiding spurious changes) and recall (capturing changes in $G^{*}$ correctly in $\hat{G}$), e.g., a segmentation probability threshold. Thus, rather than compare methods at one precision or recall level, we allow each method to expose a single parameter, and compare methods in terms of the precision-recall curve produced by varying that parameter.

To measure precision, in each No-Change scenario, we compute a binary label indicating whether the map update method performed correctly: it is correct if it does not infer any changes, i.e., if $\hat{G} = G = G^{*}$, and incorrect otherwise. We then define \emph{precision} as the fraction of No-Change scenarios where the method correctly makes no changes.

To measure recall, we use scenarios with change to compute an \emph{improvement score}, indicating the degree to which the updated road network $\hat{G}$ and post-change map $G^{*}$ are more similar than the pre-change map $G$ and $G^{*}$. Our improvement score metric can leverage any road extraction metric $S(G_1, G_2)$ that computes the similarity between two road networks $G_1$ and $G_2$. In our evaluation, we implement $S$ with either Average Path Length Similarity (APLS)~\cite{spacenet} or PixelF1~\cite{biagioni2012inferring}, which are two widely used metrics from the literature.
We define the improvement score (recall) as:
$$\text{Improvement} = \max(\frac{S(\hat{G}, G^{*}) - S(G, G^{*})}{1 - S(G, G^{*})}, 0)$$
Improvement is 0 if $\hat{G}$ offers no improvement over $G$ (e.g., $\hat{G} = G$), with $S_\text{recall}(\hat{G}, G^{*}) = S_\text{recall}(G, G^{*})$. Improvement is 1 if $\hat{G} = G^{*}$, so that $S_\text{recall}(\hat{G}, G^{*}) = 1$. We clip negative scores at 0, and average the improvement scores over the 188 scenarios with change.

Thus far, we have not considered the ambiguous roads $G^{*}_\text{extra}$: a method should be penalized neither for missing nor for inferring a road in $G^{*}_\text{extra}$. When computing precision and recall, we ignore new roads added by a method that correspond to roads in $G^{*}_\text{extra}$.

\subsection{Fusing Inferred Roads into the Map}

To evaluate prior work in road extraction on MUNO21, we propose extending these methods for map update with a post-processing step that incorporates only the parts of an inferred road network $\hat{G}$ that correspond to new roads into an existing map $G$. Our procedure is similar to prior work in map fusion~\cite{mapfuse}, and ensures that roads in $G$ that remain up-to-date are not overwritten. However, this method can only add new roads to the map, and so is only applicable to scenarios tagged Constructed or Was-Missing.

Our algorithm repeatedly finds an unprocessed edge $e_0 \in \hat{G}$ that is at least $D_\text{big} = 32$ m away from the closest edge in $G$, where $D_\text{big}$ is a distance threshold, since $e_0$ must correspond to a new road. We begin a depth-first search from $e_0$ along edges in $\hat{G}$. For each edge $(u, v)$ that we traverse during the search:
\begin{itemize}[noitemsep,topsep=0pt]
    \item If $v$ is $D$ away from edges in $G$, we add the edge $(u, v)$ to $G$, and continue the search in $\hat{G}$ from $v$.
    \item Otherwise, if we have not already explored $v$, we find the closest point $p$ to $v$ that falls along an edge $e \in G$. We split $e$ at this point, creating a new vertex $v_\text{split}$, and add an edge $(u, v_\text{split})$. We also maintain a mapping of split vertices $M$, and set $M[v] = v_\text{split}$.
    \item If we already explored $v$, we add an edge $(u, M[v])$.
\end{itemize}
We also prune components if one of two conditions hold. First, if we did not encounter any vertex $v$ close to $G$ during the search, then we discard the component, since roads that are not connected to the existing road network are not useful and likely spurious. Second, if the total length of the component is less than 50 m, we discard the component; this heuristic mitigates noisy inferred segments.

\section{Leveraging MUNO21 Training Scenarios} \label{sec:methods}

In this section, we propose a set of map update methods that leverage the tag and construction year annotations in scenarios in the MUNO21 training set, as well as aerial image time series data, to either reduce error rates or generalize to Deconstructed and Was-Incorrect scenarios. These methods would not be possible without the new types of training data that MUNO21 provides.

In particular, we consider four tasks: removing incorrect roads, pruning spurious inferred roads, pruning inferred roads that do not correspond to recent physical construction, and removing de-constructed roads. We formulate all four tasks as binary road segment classification tasks, where we train a model to classify segments as positive or negative under task-specific binary labels. We define a road segment as a sequence of edges that begins and ends at endpoint vertices, where an endpoint is any vertex with exactly one edge (a dead-end) or more than two edges (a junction); thus, a road segment is simply an uninterrupted stretch of road.

We first detail the model architecture that we employ to implement all four tasks. We then describe each task.

\begin{figure}[t]
\begin{center}
	\includegraphics[width=0.84\linewidth]{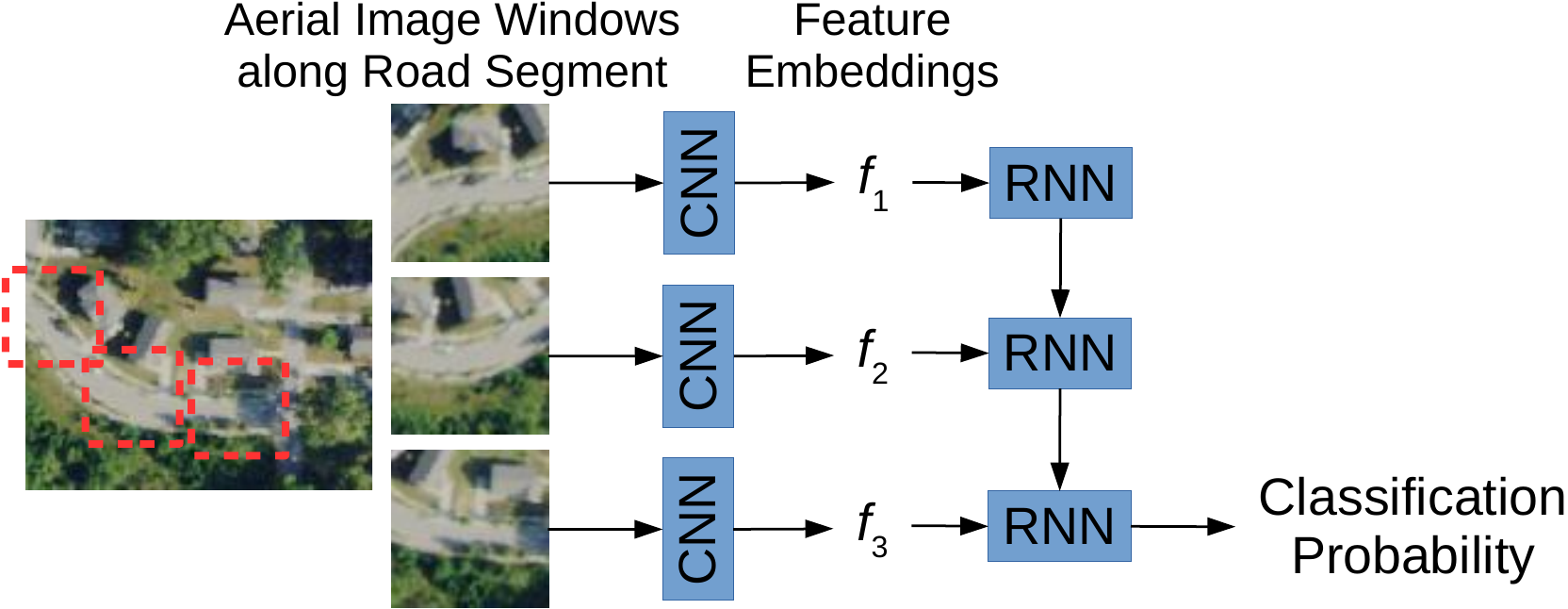}
\end{center}
	\caption{Recurrent CNN model. RNN computes probability from feature embeddings extracted by CNN.}
\label{fig:model}
\end{figure}

\smallskip
\noindent
\textbf{Recurrent CNN Model.} We employ a recurrent CNN model (Figure \ref{fig:model}) for high-accuracy road segment classification. We collect several small aerial image windows along a road segment, apply a series of convolutional layers to extract features at each window, and then apply an RNN over the feature sequence to produce a final binary classification probability through sigmoid activation on the last recurrent step. We rotate the image windows so that road direction is aligned with the image x axis. This model naturally handles road segments of very different lengths.

\smallskip
\noindent
\textbf{Removing Incorrect Roads (Remove-Incorrect).} First, we can use the Deconstructed and Was-Incorrect training scenarios to train the model to classify incorrect roads in the base map $G$. We use each de-constructed or incorrect road segment in the scenarios as a positive example. We use every segment in the base map of No-Change scenarios as negative examples. We train the model to classify incorrect segments given only the most recent aerial image. On test scenarios, we produce an inferred road network $\hat{G}$ by pruning segments in $G$ where the model estimates the segment to be incorrect with probability greater than a threshold.

\smallskip
\noindent
\textbf{Pruning Spurious Inferred Roads (Prune-Spurious).} We can also apply our model to prune spurious road segments in a road network $\hat{G}$ inferred by a road extraction method: we train the model to predict whether an inferred road is correct (matches a road in $G^{*}$) or spurious. Our approach is similar to the connection classifier in DeepRoadMapper~\cite{mattyus2017deeproadmapper}.

To train the classifier, we obtain positive and negative examples by applying a road extraction method in every scenario in the MUNO21 training set, and heuristically determining if each inferred road is correct or spurious:
we say a segment is correct if every point on the segment is within $D_\text{small} = 8$ m of some edge in $G^{*}$, and spurious if there is any point on the segment $>D_\text{big} = 32$ m away from $G^{*}$.

We apply the model on test scenarios as follows: we first apply the road extraction method to compute an initial inferred road network $\hat{G}$. We then prune segments in $\hat{G}$ where the model has high confidence that the segment is spurious. Finally, we fuse the remaining segments with $G$.

\smallskip
\noindent
\textbf{Focusing on Constructed Roads (Focus-Constructed).} Thus far, we have focused on map update methods that input only the most recent aerial image. However, as we will show in the evaluation, these methods have limited accuracy, and can yield high error rates. One way to reduce errors is to compare aerial images over time, and only update the map when we observe a change in the physical road network visible in the imagery. Although this approach can only handle Constructed and Deconstructed scenarios (where there are actual changes to the physical road network), focusing on these scenarios may be necessary in practice to achieve error rates suitable for full automation.

For newly constructed roads, we can focus on them by pruning roads inferred by a road extraction method that do not seem to correspond to recent physical construction. For this task, the model inputs windows of two aerial images captured at different times instead of just one; thus, the CNN processes a 6-channel input. We obtain positive examples of construction from the scenarios tagged Constructed in the MUNO21 training set, and negative examples from scenarios tagged No-Change. We train the model to predict whether the second input image contains new roads that are not visible in the first image. We also use De-constructed road segments as positive examples, but reverse the order of the aerial images in the input.

The inference procedure is similar to Prune-Spurious: after applying a road extraction method, we process each inferred segment through the model, and prune segments where the model has low confidence that there is a new road.

\smallskip
\noindent
\textbf{Focus-Deconstructed.} We reuse the Focus-Constructed model to prune de-constructed roads from a base map.
Our inference procedure is similar to Remove-Incorrect: we remove each segment in $G$ where the model has high confidence that the segment was present in the earliest available aerial image but no longer exists in the most recent image.

\section{Evaluation} \label{sec:eval}

We now evaluate five state-of-the-art road extraction and map update methods, along with the extensions that we have proposed, on the MUNO21 test scenarios.

\smallskip
\noindent
\textbf{Methods.} We evaluate four road extraction methods, which we extend for map update using our map fusing approach: RoadTracer~\cite{roadtracer}, RecurrentUNet~\cite{yang2019road}, Sat2Graph~\cite{sat2graph}, and RoadConnect~\cite{batra2019improved}.
Additionally, we evaluate the map update method MAiD~\cite{maid}, which is able to add new roads to an existing map but does not address other challenges in updating maps.
Lastly, we evaluate the extensions we proposed in Section \ref{sec:methods} in conjunction with RoadTracer. We train all methods on 3000 sq km of aerial image and map data from the 10 cities in the MUNO21 training set.

\smallskip
\noindent
\textbf{Metrics.} We evaluate the methods over the 568 map update scenarios in the 11-city MUNO21 test set on their precision-recall curves.
We obtain curves by varying one parameter in each method that provides a tradeoff between precision and recall.
We derive recall from Average Path Length Similarity (APLS)~\cite{spacenet} improvement scores as discussed in Section \ref{sec:dataset}; these scores quantify how much an inferred road network $\hat{G}$ improves over the pre-change map $G$. APLS computes shortest paths between several pairs of corresponding points in $G^{*}$ and $\hat{G}$, and compares their lengths. We also compare curves where recall is derived from PixelF1~\cite{biagioni2012inferring} improvement in the supplementary material.

\begin{figure*}[t]
    \centering
    \includegraphics[width=\linewidth]{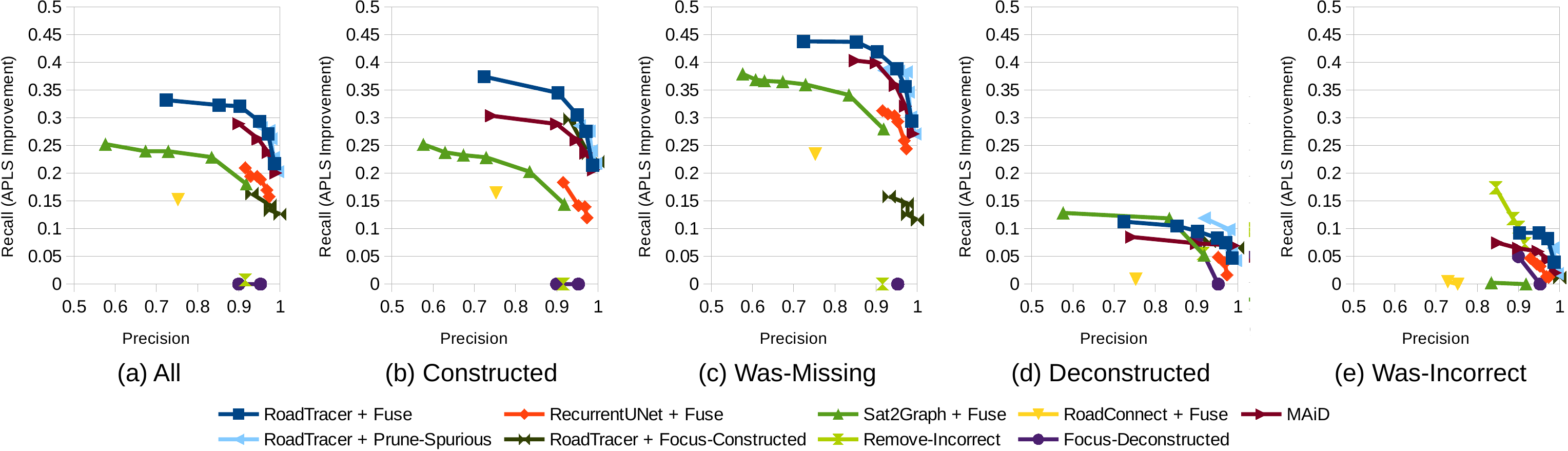}
    \caption{Precision-recall curves on the MUNO21 test set. Recall is measured in terms of APLS Improvement, which quantifies how much better an inferred map $\hat{G}$ captures the post-change map $G^{*}$ than the pre-change map $G$. In addition to reporting curves averaged over all scenarios with change in (a), we break it down by tags in (b) through (e).}
    \label{fig:results}
\end{figure*}

\begin{figure*}[t]
\begin{subfigure}{\linewidth}
    \centering
    \setlength{\tabcolsep}{1pt}
    \begin{tabular}{cccccccc}
        & 2012 & 2019 & Ground Truth & RoadTracer & RecurrentUNet & Sat2Graph & MAiD \\
        
        \rotatebox{90}{~~~Constructed} &
    	\includegraphics[width=0.13\linewidth]{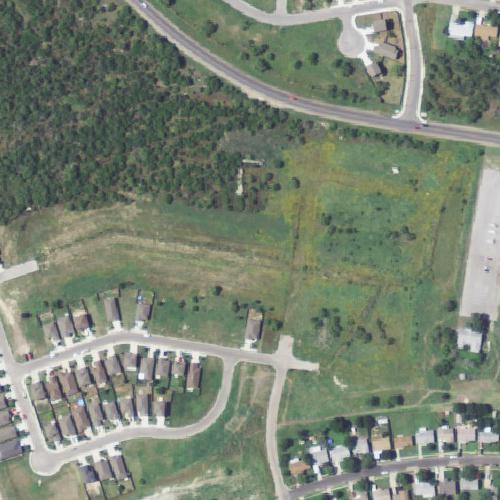} &
    	\includegraphics[width=0.13\linewidth]{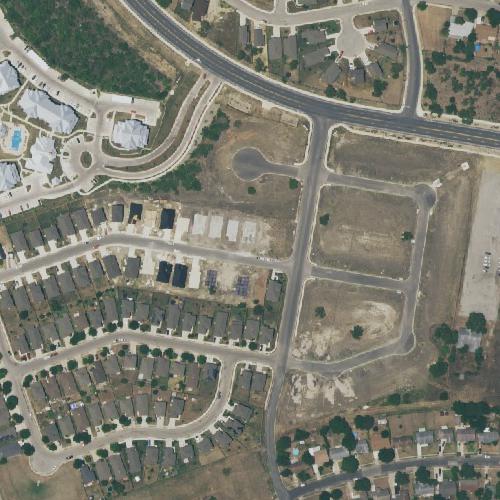} &
    	\includegraphics[width=0.13\linewidth]{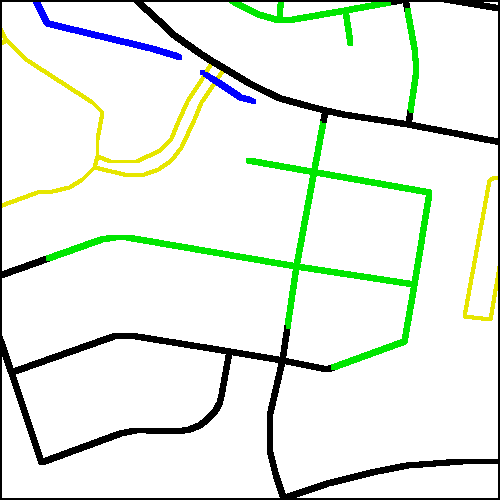} &
    	\includegraphics[width=0.13\linewidth]{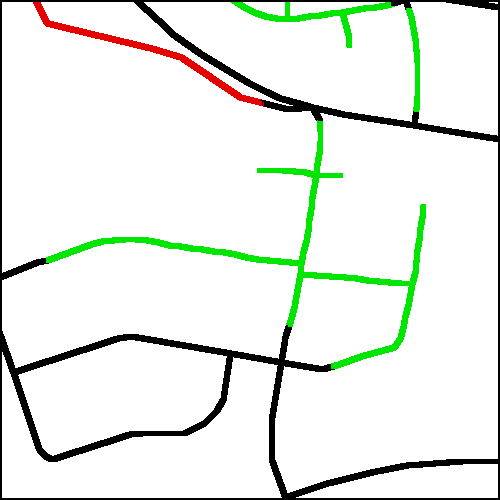} &
    	\includegraphics[width=0.13\linewidth]{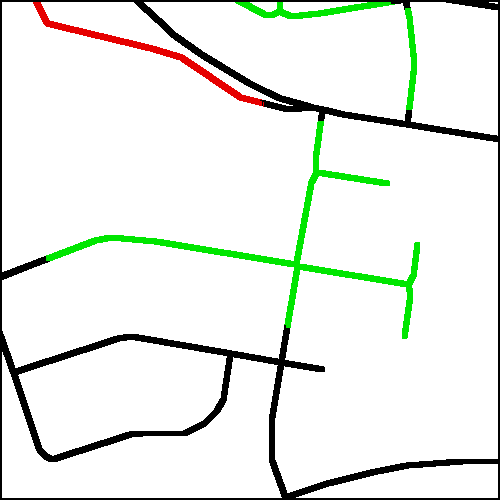} &
    	\includegraphics[width=0.13\linewidth]{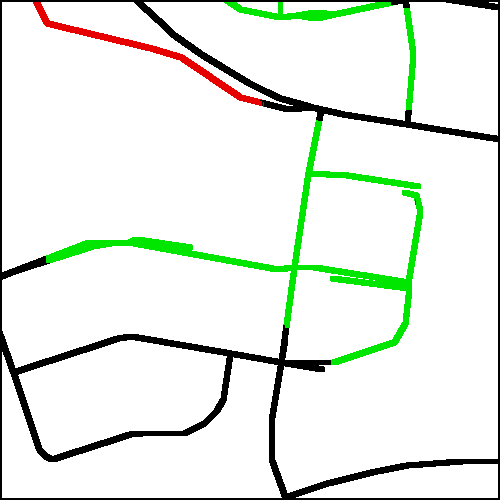} &
    	\includegraphics[width=0.13\linewidth]{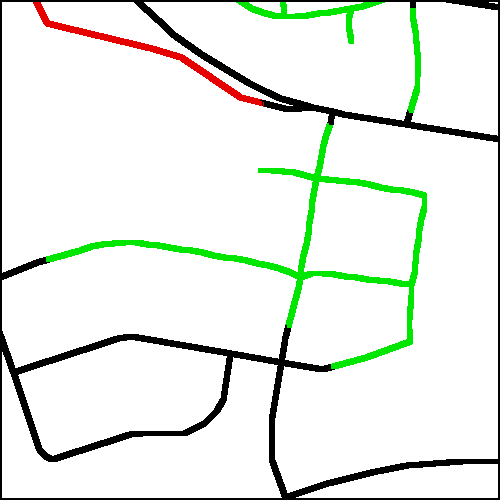} \\
        
        \rotatebox{90}{~~Was-Missing} &
    	\includegraphics[width=0.13\linewidth]{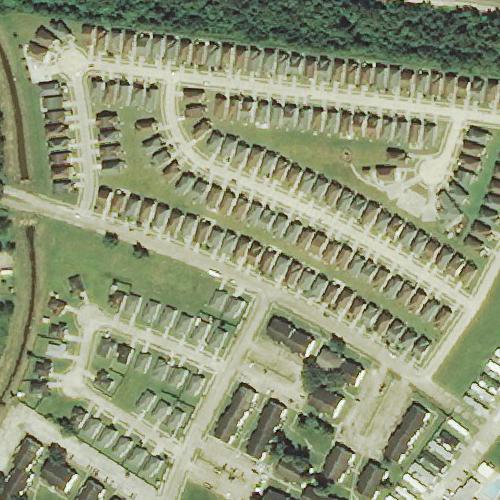} &
    	\includegraphics[width=0.13\linewidth]{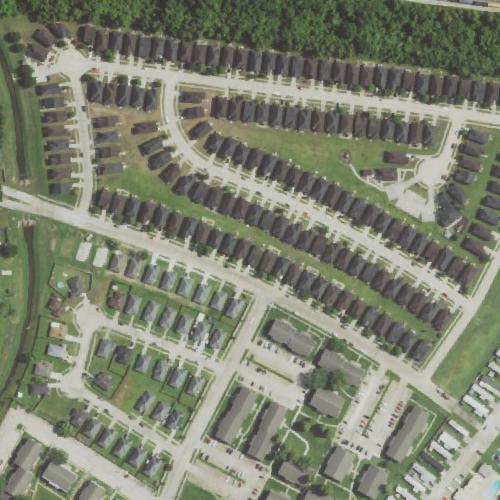} &
    	\includegraphics[width=0.13\linewidth]{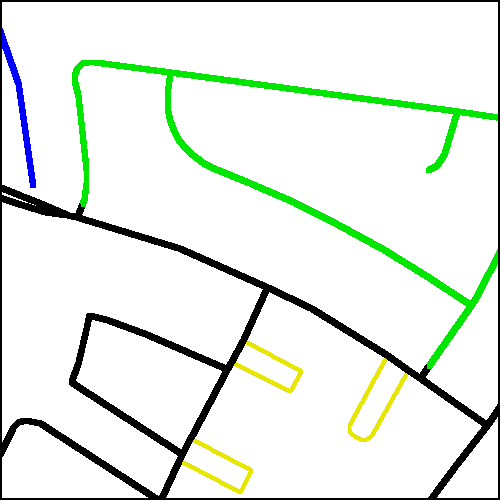} &
    	\includegraphics[width=0.13\linewidth]{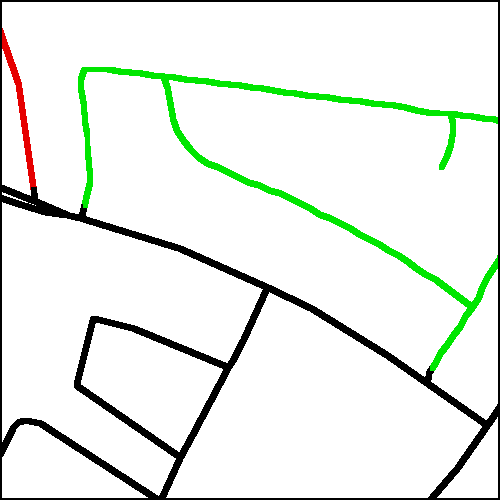} &
    	\includegraphics[width=0.13\linewidth]{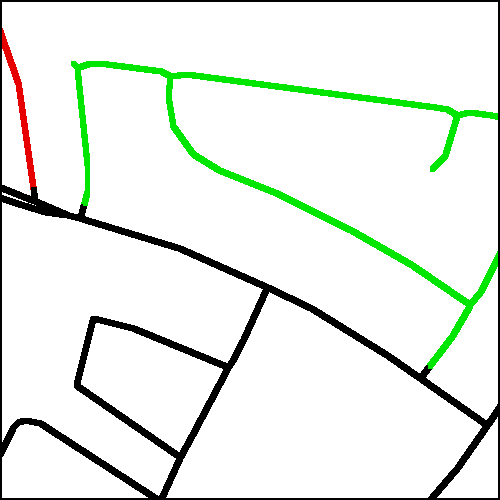} &
    	\includegraphics[width=0.13\linewidth]{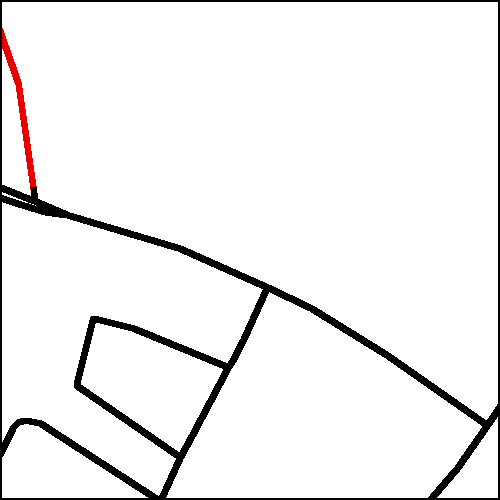} &
    	\includegraphics[width=0.13\linewidth]{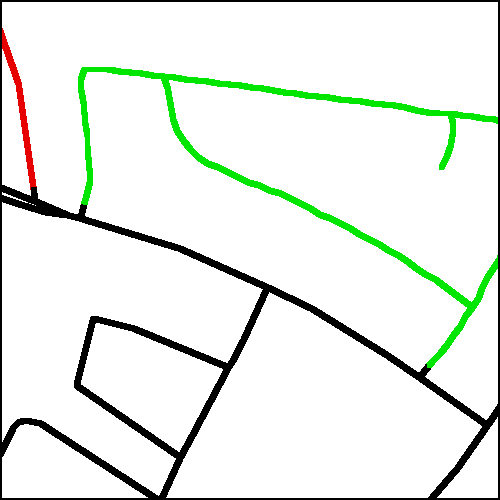} \\
        
        \rotatebox{90}{De/Constructed} &
    	\includegraphics[width=0.13\linewidth]{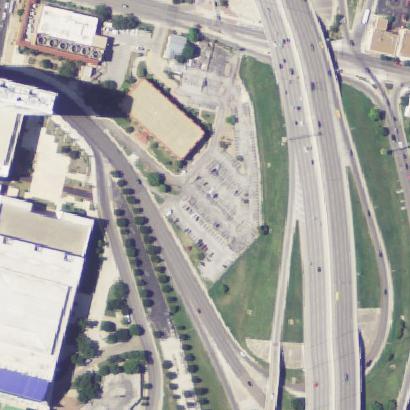} &
    	\includegraphics[width=0.13\linewidth]{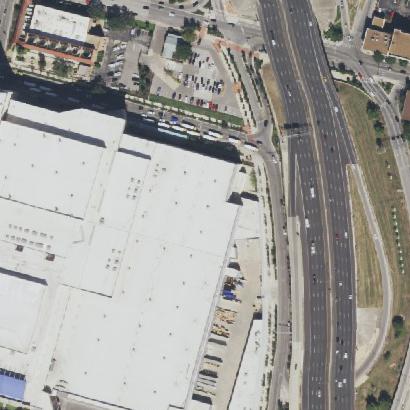} &
    	\includegraphics[width=0.13\linewidth]{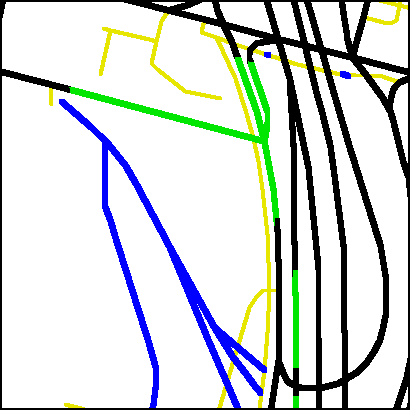} &
    	\includegraphics[width=0.13\linewidth]{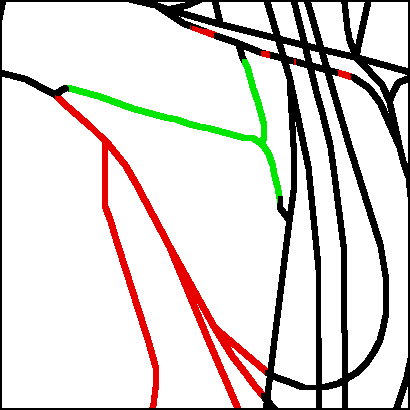} &
    	\includegraphics[width=0.13\linewidth]{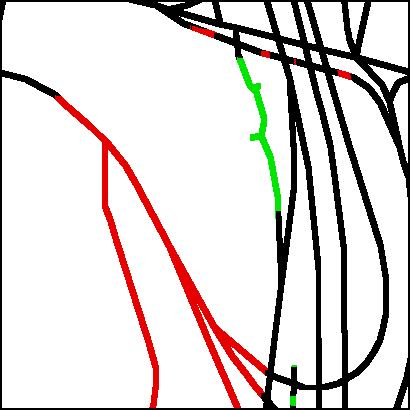} &
    	\includegraphics[width=0.13\linewidth]{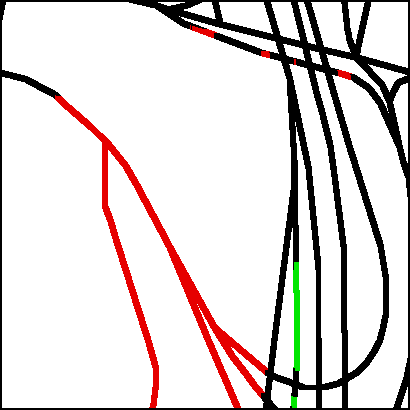} &
    	\includegraphics[width=0.13\linewidth]{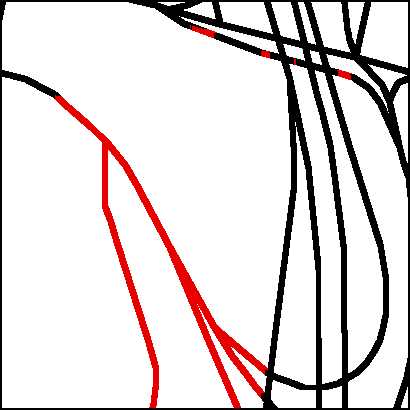} \\
	\end{tabular}
	\vspace{-7pt}
	\caption{Outputs of road extraction methods, with map fusing. These methods are only able to add new roads to the map.}
\end{subfigure}

\begin{subfigure}{0.38\linewidth}
    \centering
    \setlength{\tabcolsep}{1pt}
    \begin{tabular}{cc}
        2019 & Remove-Incorrect \\
        \includegraphics[width=0.34\linewidth]{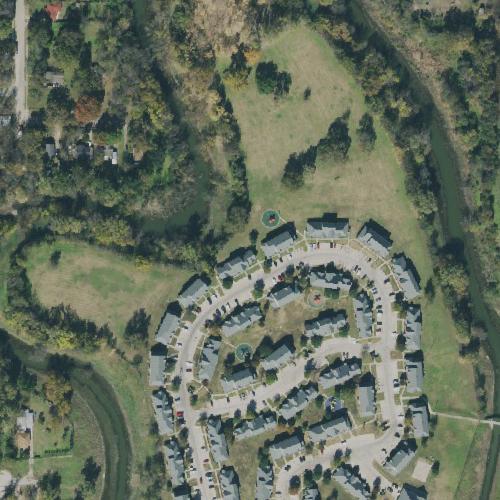} &
        \includegraphics[width=0.34\linewidth]{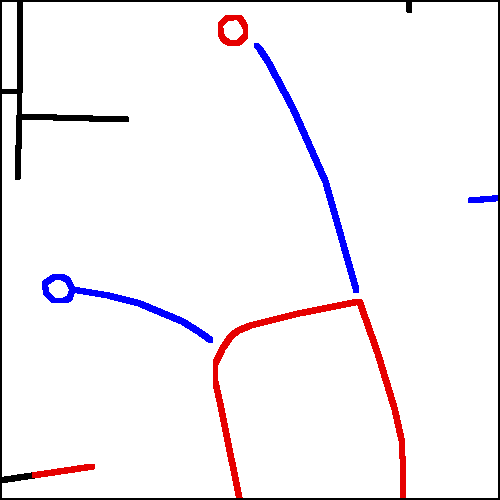} \\
    \end{tabular}
	\vspace{-4pt}
    \caption{An example with Remove-Incorrect.}
\end{subfigure}
\begin{subfigure}{0.58\linewidth}
    \centering
    \setlength{\tabcolsep}{1pt}
    \begin{tabular}{cccc}
        2019 & Ground Truth & MAiD & StreetView \\
        \includegraphics[width=0.22\linewidth]{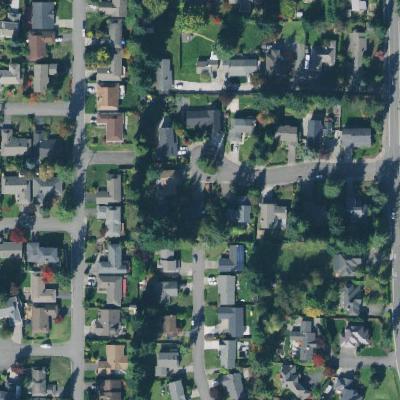} &
        \includegraphics[width=0.22\linewidth]{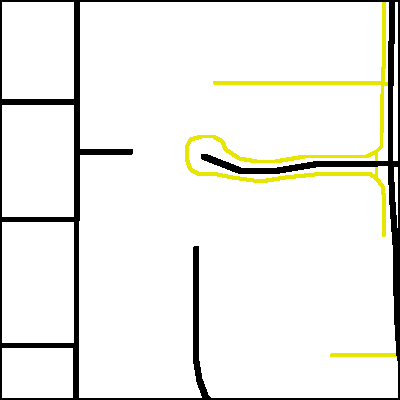} &
        \includegraphics[width=0.22\linewidth]{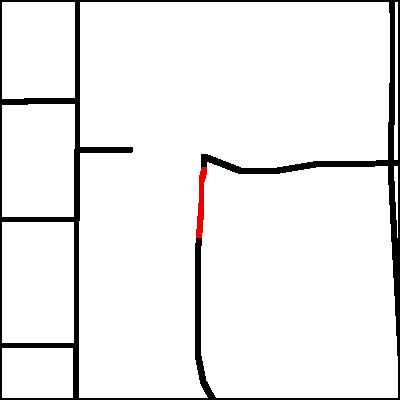} &
        \includegraphics[width=0.22\linewidth]{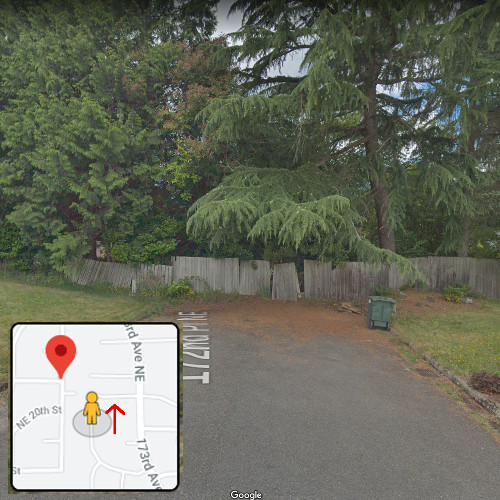} \\
    \end{tabular}
	\vspace{-4pt}
    \caption{An example error: a spurious connection is inferred.}
\end{subfigure}
\caption{Qualitative results on diverse map update scenarios. We show added roads in green (correct) or red (wrong), and removed roads in blue (correct) or orange (wrong). The base map $G$ is black, and edges in $G_\text{extra}$ are yellow.}
\label{fig:qual}
\end{figure*}

\smallskip
\noindent
\textbf{Results.} We show quantitative results in terms of precision-recall curves in Figure \ref{fig:results}, and qualitative results in Figure \ref{fig:qual}.

Among the five prior methods, at 90\% precision, RoadTracer provides 32\% recall, while Sat2Graph provides 18\% recall, indicating that the former better captures road topology. However, even when we consider only Constructed and Was-Missing scenarios (since these methods do not handle road removal), none of the methods provide recall greater than 45\%. Furthermore, at 30\% recall, no method provides precision above 95\%, implying a 5\% error rate.

To improve precision, we can extend a method with our Prune-Spurious and Focus-Constructed approaches proposed in Section \ref{sec:methods}. We evaluate these approaches with RoadTracer. Prune-Spurious increases precision up to 99.5\% at 20\% recall, but drops in recall yield a curve that is comparable to that of RoadTracer alone. On scenarios tagged Constructed in Figure \ref{fig:results}(b), Focus-Constructed further improves precision to 99.8\% at 20\% recall; however, because this method only adds roads that correspond to recent physical construction, it does not perform well over the entire dataset.

We also evaluate our Remove-Incorrect and Focus-Deconstructed road removal methods. Remove-Incorrect provides up to 17\% recall on Was-Incorrect scenarios in Figure \ref{fig:results}(e), suggesting that it succeeds to a limited extent at removing incorrect roads from the map. However, its performance is limited because it often removes incorrect roads while leaving intact neighboring incorrect roads, which creates disconnections that APLS harshly penalizes; for example, the red cul-de-sac at the top of Figure \ref{fig:qual}(b) is retained in the map but disconnected from other roads.

Focus-Deconstructed performs poorly, even on Deconstructed scenarios in Figure \ref{fig:results}(d). We find that this is because the classifier is trained with weak negative examples (unchanged roads) where there are no road network changes in the surroundings. Then, in Deconstructed scenarios, it removes not only the deconstructed roads, but nearby unchanged roads as well.

Overall, we argue that substantial further improvements are needed to realize automatic map update. Map update methods should achieve precision greater than 99\% to be useful in practice: otherwise, they would produce an enormous number of errors along the vast majority of the existing map that is up-to-date --- 99\% precision in the US implies that 12,000+ errors\footnote{We multiply the 1\% error rate by 12M km of roads in the US, and divide by 10 km of roads on average in each no-change scenario.} would be introduced on each automatic update.
To further validate this, we show an example error introduced by MAiD in Figure \ref{fig:qual}(c): MAiD adds a spurious connection between two disconnected segments. This is a severe error: in fact, similar errors made by an industry mapping team exploring semi-automation in editing maps garnered widespread criticism from the OSM Thailand community, forcing the team to increase the training of map validators\footnote{See \url{https://forum.openstreetmap.org/viewtopic.php?pid=684197\#p684197}. Regarding a spurious inferred connection, one user wrote: drivers ``are going to get a tad [annoyed] being routed down there now and having to do a U-turn when they get to the fence''.}. Indeed, many OSM contributors argue that spurious roads are far more problematic than missing roads\footnote{See \url{https://forum.openstreetmap.org/viewtopic.php?pid=712005\#p712005}.}. Thus, minimizing errors is crucial in automatic map update --- and much work remains to be done, since the best tested method yields just 95\% precision at 30\% recall.

\section{Conclusion}

Automatic methods for extracting road networks from GPS trajectories and aerial images have progressed greatly in the last decade. However, for these methods to be useful in practice, they must be applied to update existing maps. MUNO21 is a large-scale and comprehensive dataset for map update using aerial images that serves as a substantially better benchmark for the practical deployment of automatic map inference methods than prior datasets. MUNO21 consists of over a thousand map update scenarios, which exhibit diverse and complex challenges including adding new roads, removing incorrect roads, and updating maps to reflect physical road topology changes. We believe that MUNO21 will help the community transition from road extraction to the map update task.

{\small
\bibliographystyle{ieee_fullname}
\bibliography{egbib}
}

\end{document}